\documentclass{article}

\usepackage{arxiv}

\usepackage[utf8]{inputenc} 
\usepackage[T1]{fontenc}    
\usepackage{hyperref}       
\usepackage{url}            
\usepackage{booktabs}       
\usepackage{amsfonts}       
\usepackage{nicefrac}       
\usepackage{microtype}      
\usepackage{lipsum}		
\usepackage{graphicx}
\usepackage{natbib}
\usepackage{doi}
\usepackage{enumitem}
\usepackage{float}
\usepackage{graphicx}
\usepackage{graphicx}
\usepackage{subcaption}
\usepackage{tabularx}
\usepackage[most]{tcolorbox} 

\newlength{\BlockImgH}
\newlength{\BlockGap}
\newlength{\BlockBoxH}
\usepackage[most]{tcolorbox}
\tcbuselibrary{listings,breakable}
\usepackage{listings}

\lstdefinestyle{promptstyle}{
  basicstyle=\ttfamily\scriptsize,
  columns=fullflexible,
  breaklines=true,
  breakatwhitespace=true,
  showstringspaces=false
}

\newtcblisting{promptlisting}[2][]{%
  breakable,
  enhanced,
  title={#2},
  colback=white,
  colframe=black!35,
  boxrule=0.3pt,
  left=1mm,right=1mm,top=1mm,bottom=1mm,
  listing only,
  listing options={style=promptstyle},
  #1
}

\newtcolorbox{promptbox}[1][]{%
  boxrule=0.3pt,
  colback=white,
  colframe=black!35,
  left=1mm,right=1mm,top=1mm,bottom=1mm,
  before upper={\setlength{\parskip}{0pt}\setlength{\parindent}{0pt}},
  fontupper=\ttfamily\scriptsize,
  breakable,
  title=#1
}
\usepackage{graphicx}
\title{\texorpdfstring{%
  \raisebox{-0.25\height}{\includegraphics[height=0.8cm]{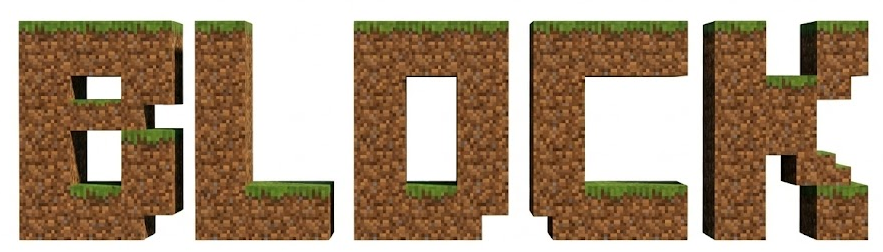}}%
}{BLOCK}: An Open-Source Bi-Stage MLLM Character-to-Skin Pipeline for Minecraft}

\author{
  Hengquan~Guo\\
  ShanghaiTech University\\
  \texttt{guohq@shanghaitech.edu.cn}\\
}

\date{}

\renewcommand{\headeright}{Technical Report}
\renewcommand{\undertitle}{Generating pixel-perfect Minecraft skins from any character concept.}
\renewcommand{\shorttitle}{BLOCK (v0.5).}

\hypersetup{
  pdftitle={BLOCK: An Open-Source Bi-Stage MLLM Character-to-Skin Pipeline For Minecraft},
  pdfsubject={Technical Report},
  pdfauthor={Hengquan Guo, ShanghaiTech University},
  pdfkeywords={Minecraft skin, MLLM, character rendering, UV atlas, pipeline},
}

\begin{document}
\maketitle

\begin{abstract}
We present \textbf{BLOCK}\footnote{https://huggingface.co/AliceKJ/BLOCKv0.5.}, an open-source bi-stage character-to-skin pipeline that generates pixel-perfect Minecraft skins from arbitrary character concepts. BLOCK decomposes the problem into (i) a \textbf{3D preview synthesis stage} driven by a large multimodal model (MLLM) with a carefully designed prompt-and-reference template, producing a consistent dual-panel (front/back) oblique-view Minecraft-style preview; and (ii) a \textbf{skin decoding stage} based on a fine-tuned FLUX.2 model that translates the preview into a skin atlas image. We further propose \textbf{EvolveLoRA}, a progressive LoRA curriculum (text-to-image $\rightarrow$ image-to-image $\rightarrow$ preview-to-skin) that initializes each phase from the previous adapter to improve stability and efficiency. BLOCK is released with all prompt templates and fine-tuned weights to support reproducible character-to-skin generation.

\end{abstract}


\section{Introduction}

\begin{figure}[h]
    \centering
    \includegraphics[width=0.9\linewidth]{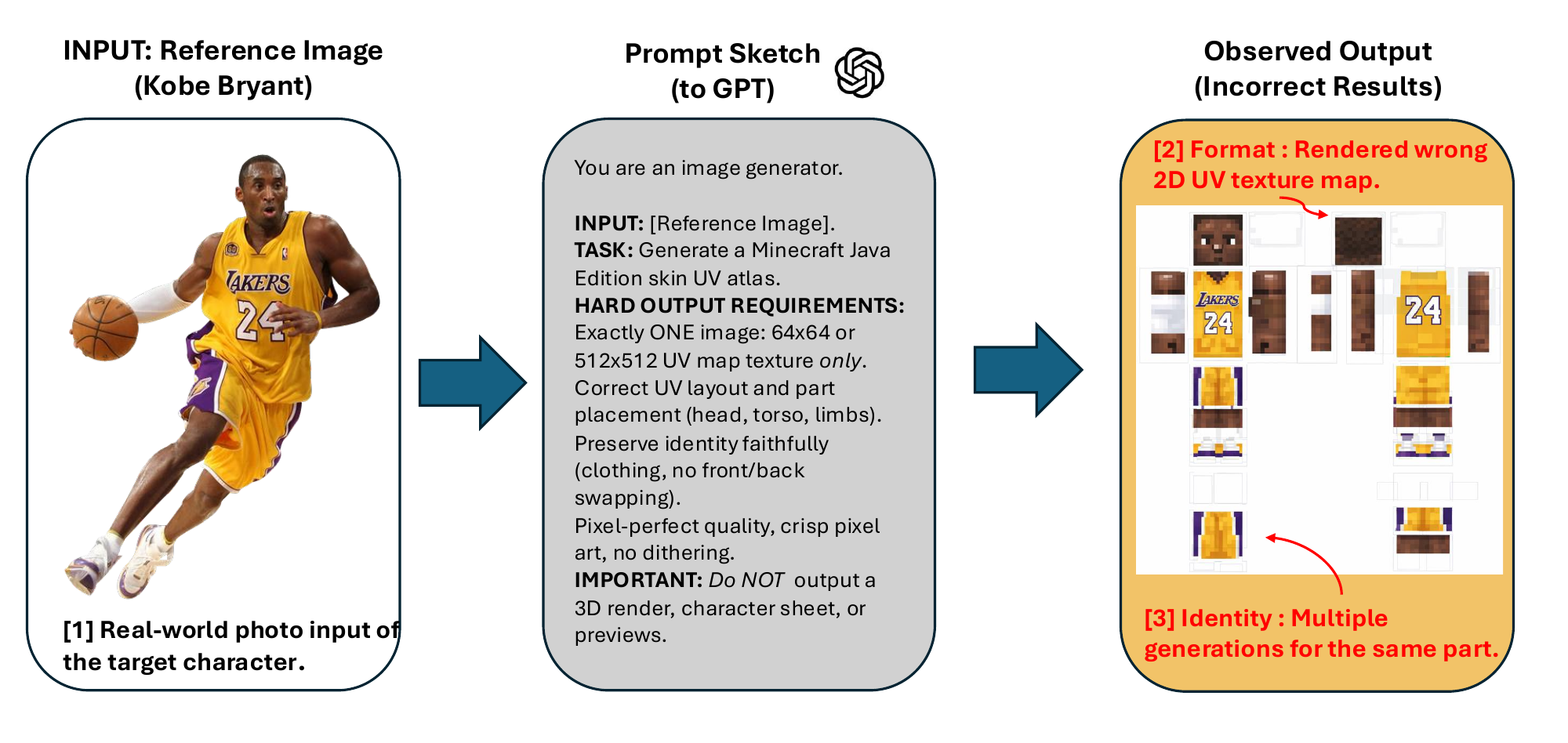}
    \caption{Current Powerful MLLMs fail to generate Minecraft skin from character concept.}
    \label{fig:introduction}
\end{figure}

Minecraft skins are lightweight 64$\times$64 UV texture atlases, yet generating \emph{valid} and \emph{pixel-perfect} skins from arbitrary character concepts remains surprisingly difficult (Figure \ref{fig:introduction}). Modern multimodal LLMs (MLLMs) already exhibit a strong base capability in understanding character references and following detailed layout instructions. However, due to a mismatch between their training data and the Minecraft skin domain, one-stage generation towards the current powerful MLLMs is often brittle: it is hard to compress and align high-level character traits into the correct pixel blocks and UV regions; and (ii) the generated textures frequently violate strict skin constraints. The challenge is therefore not merely producing plausible pixel art, but producing a structurally correct UV atlas with correct part placement, clean boundaries, and consistent design across base and overlay layers. Meanwhile, real users typically begin with character images rather than UV atlases, making direct UV generation even more error-prone.

To make Minecraft character-to-skin generation robust under strict UV constraints, we propose \textbf{BLOCK}, a practical open-source pipeline that separates the problem into two stages:
\begin{itemize}[leftmargin=*,itemsep=0.2em,topsep=0.2em]
    \item \textbf{Stage-1 (Character $\rightarrow$ 3D Preview)}: a large multimodal model consumes the user-provided character reference and, using a fixed prompt template with reference images, synthesizes a Minecraft-style dual-panel preview with \textbf{front/back} views. In this part, we use Gemini Nano Banana Pro, which is the only model we tested that consistently succeeds on this task.
  \item \textbf{Stage-2 (Preview $\rightarrow$ Skin Atlas)}: a fine-tuned FLUX.2 model performs image-conditioned generation to produce a \textbf{512$\times$512} skin atlas. A deterministic decoder then converts the output into a valid \textbf{64$\times$64} skin, including overlay regions.
\end{itemize}

This design is motivated by the observation that MLLMs are robust at interpreting character references and obeying layout/style constraints to produce coherent Minecraft-like previews, while a specialized diffusion/flow model can be trained to solve the UV-structured translation task with high fidelity. BLOCK thus avoids asking a single model to simultaneously solve semantic understanding, 3D rendering, and strict UV atlas validity.

The fine-tuning phase is based on the open-source dataset, \href{https://huggingface.co/datasets/nyuuzyou/Minecraft-Skins-20M}{\texttt{Minecraft-Skins-20M}}, which contains about 200K unique Minecraft skins data. We employ a three-phase fine-tuning curriculum.
In Phase I, we perform text-to-image training only, where the text prompt is an automatically constructed description inferred from the skin itself (e.g., coarse colors and patterns of major body parts).
In Phase II, we switch to image-to-image training, using paired \textbf{front/back} renders of the same skin as conditioning images.
In Phase III, we train the final preview-to-skin model that translates a Minecraft-style 3D dual-panel preview into a UV skin atlas.
Each phase initializes its LoRA weights from the adapter obtained in the previous phase rather than starting from scratch.
We refer to this progressive adapter initialization strategy as \textbf{EvolveLoRA}. In summary, in this report,
\begin{itemize}[leftmargin=*,itemsep=0.2em,topsep=0.2em]
  \item We present \textbf{BLOCK}, an open-source \emph{two-phase} character-to-skin pipeline that maps arbitrary character references to Minecraft skins.
  \item We design a scalable data specification and pair-construction pipeline that converts a large skin corpus into supervised image-conditioned training data (front/back renders and 3D-style previews) for UV atlas synthesis.
  \item We propose EvolveLoRA, a progressive LoRA initialization curriculum that reuses the adapter from the previous phase (text-to-image $\rightarrow$ image-to-image $\rightarrow$ preview-to-skin) to improve training stability and efficiency.
  \item We release the complete implementation, prompt templates to enable reproducible evaluation and easy extension.
\end{itemize}

\section{Related Work}
\label{sec:related}

\paragraph{Parameter-Efficient Fine-Tuning (PEFT) and LoRA variants.}
Low-Rank Adaptation (LoRA) \cite{hu2022lora} is a widely adopted parameter-efficient fine-tuning method that injects trainable low-rank matrices into selected linear projections, enabling efficient adaptation while keeping most base weights frozen. Subsequent work explored PEFT under various objectives. QLoRA \cite{dettmers2023qlora} combines low-rank adaptation with quantization to reduce memory footprint while preserving fine-tuning quality. AdaLoRA \cite{zhang2023adalora} dynamically allocates rank budgets across layers/modules, aiming to concentrate capacity where it matters most. DoRA \cite{liu2024dora} proposes a weight-decomposed formulation to improve adaptation behavior compared to conventional LoRA parameterization.

\paragraph{Continual adaptation and repeated low-rank updates.}
A recurring challenge in PEFT is how to efficiently accumulate new capabilities without catastrophic interference. ReLoRA \cite{lialin2023relora} studies training regimes that repeatedly apply low-rank updates while periodically ``resetting'' or reparameterizing, aiming to recover benefits of higher-rank training while maintaining low-rank efficiency. Our \textbf{EvolveLoRA} is inspired by this line: rather than training a single adapter from scratch for the final target, we warm-start new adapters from previously learned LoRA weights and evolve the task difficulty.

\section{System Overview}

\begin{figure}[H]
    \centering
    \includegraphics[width=1\linewidth]{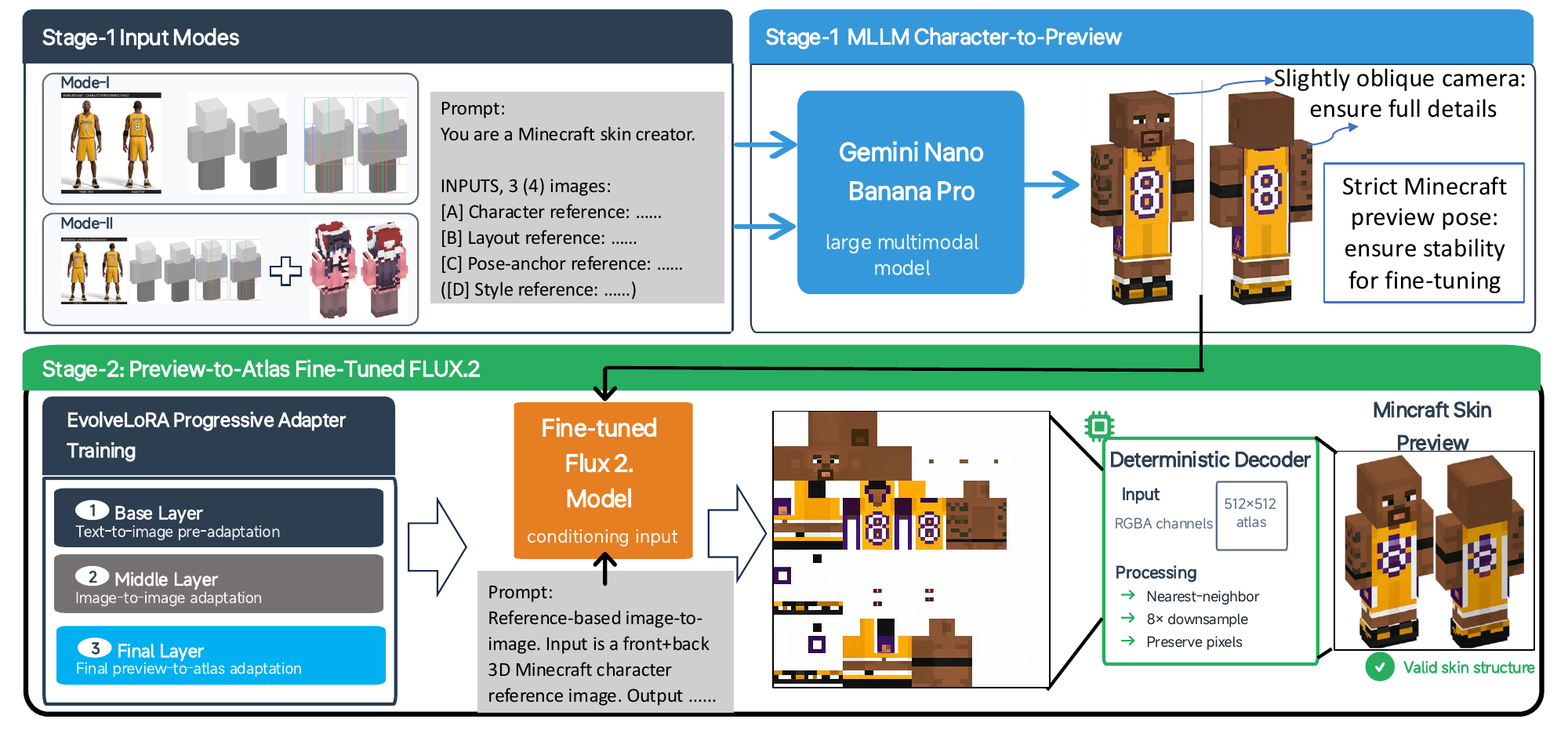}
    \caption{Overview of BLOCK.}
    \label{fig:pipeline}
\end{figure}

Figure~\ref{fig:pipeline} illustrates the BLOCK pipeline. Stage-1 provides two input modes: in Mode-I, the MLLM takes three images---a user character reference with front/back views (A), a layout anchor (B) that fixes the dual-panel framing, and a pose anchor (C) that enforces a strict Minecraft standing pose; in Mode-II, an additional style reference (D) is included (a Minecraft-skin 3D preview) to control the rendering style. In both cases, we use \textbf{Gemini Nano Banana Pro} to synthesize a consistent dual-panel Minecraft-style preview under a slightly oblique camera, which exposes richer details while keeping the pose stable for downstream learning. The Stage-1 preview then serves as the conditioning input to Stage-2, where a \textbf{fine-tuned FLUX.2} model translates the preview into a $512\times512$ skin atlas. Fine-tuning follows \textbf{EvolveLoRA}, a progressive adapter curriculum that evolves from text-to-image, to front/back-conditioned image-to-image, and finally to preview-to-atlas adaptation. Finally, a deterministic decoder converts the atlas into a valid $64\times64$ RGBA skin by performing an $8\times$ downsampling and enforcing the Minecraft skin structure, yielding a ready-to-use skin preview.

\subsection{Stage-1: MLLM Character-to-Preview}

Stage-1 uses \textbf{Gemini Nano Banana Pro} as the large multimodal model (the only model we tested that consistently succeeds on this task). We support two input modes built on a fixed prompt template:

\textbf{Mode-I (3-image template).} The model receives three images: (A) the user-provided character reference containing \textbf{front/back} views; and (B,C) two \textbf{layout/template anchors} that specify the target dual-panel framing (front on the left, back on the right), fixed margins, and a consistent character scale. This mode focuses on reliable composition and view correctness.

\textbf{Mode-II (3+1 images with style reference).} Based on Mode-I, we add an optional fourth image as a \textbf{style anchor}: a \emph{Minecraft skin 3D preview} rendered in the desired target style. This reference controls whether the synthesized preview should appear as anime-style, realistic, or other stylizations, while keeping the character identity tied to the user reference.

\textbf{Mode-III (meta-prompt injection from a single character reference).}[Recommended]
In this mode, the model takes only (A) a user-provided character concept/reference image and a fixed \textbf{meta-prompt}. An MLLM first extracts salient character attributes from the image (e.g., hairstyle structure, dominant colors, outfit components, and key accessories) and \emph{injects} them into the meta-prompt as a structured, minimal attribute block. The meta-prompt itself is designed to emphasize \textbf{Minecraft-specific anchors} and \textbf{format constraints} (dual-panel front/back layout, camera, pose, background, and pixel-rendering cues). Unlike Mode-I/II, Mode-III does not rely on long hand-written natural-language constraints to preserve character fidelity; instead, character-specific details are carried by the injected attribute block, while the meta-prompt remains largely invariant across users. This yields a more modular prompting interface: layout/format control is fixed and reusable, and character identity is provided through automatic prompt injection rather than manual prompt engineering.

\begin{figure*}[t]
\centering

\begin{minipage}[t]{0.40\textwidth}
\vspace{0pt}
\includegraphics[width=\linewidth,height=\BlockImgH,keepaspectratio]{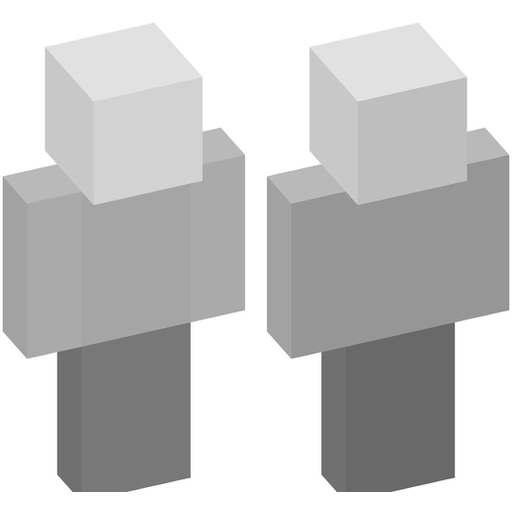}\\[\BlockGap]
\includegraphics[width=\linewidth,height=\BlockImgH,keepaspectratio]{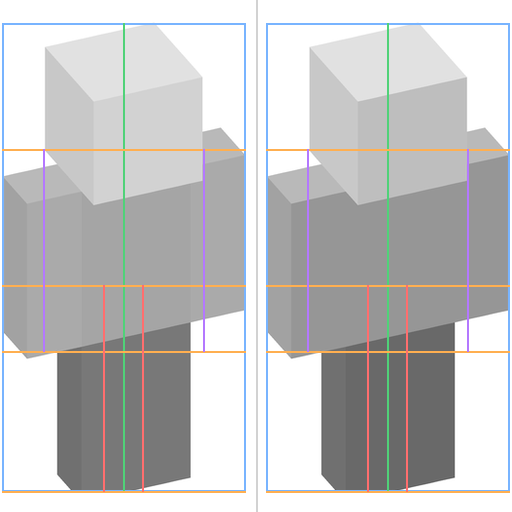}
\end{minipage}\hfill
\begin{minipage}[t]{0.58\textwidth}
\vspace{0pt}
\begin{tcolorbox}[
  enhanced,
  title={Mode-I (3-image template) prompt excerpt},
  height=\BlockBoxH,
  valign=top,
  sharp corners,
  boxrule=0.3pt,
  colback=white,
  colframe=black!35,
  left=1mm,right=1mm,top=1mm,bottom=1mm
]
\ttfamily\small
You are an image generator.

INPUTS (3 images):
[A] Character reference: same anime character with FRONT + BACK views (side-by-side or top/bottom).
[B] Layout reference: target Minecraft dual-panel framing and camera.
[C] Pose-anchor reference: strict limb-position guide.

OUTPUT:
Generate exactly ONE final image only.
- Pure white background \#FFFFFF.
- Two panels only, side-by-side:
  - Left = FRONT
  - Right = BACK
- Never swap.
- No extra views, no extra images, no text, no watermark, no UI, no props, no scene.

VIEW PARSING (HARD):
- If [A] is side-by-side: default left=front, right=back (unless explicit text labels say otherwise).
- If [A] is top/bottom: rearrange to left=front, right=back.
- Left panel uses front-view info only.
- Right panel uses back-view info only.
- Do not copy front-only details to back or back-only details to front.

CAMERA / ORIENTATION (LOCKED):
- Near-orthographic look (very weak perspective).
- Front panel: +24$^\circ$ yaw (front-right), +30$^\circ$ downward tilt.
- Back panel: 180$^\circ$+24$^\circ$ yaw (back-right), same +30$^\circ$ downward tilt.
- Front/back orientation relationship must remain parallel in 3D space.
- Match framing style of [B].

POSE LOCK (ULTRA STRICT):
- Minecraft default standing pose only.
- Head, torso, arms, legs are clean cuboids with straight vertical sides.
- Arms: hang straight down; no bend/swing/spread; left/right symmetry required.
- Legs: straight down; parallel; feet share the same baseline.
- If any tolerance fails, regenerate......................................
\end{tcolorbox}
\end{minipage}

\caption{Stage-1 Mode-I input anchors (B/C) and a prompt excerpt (Gemini Nano Banana Pro).}
\label{fig:stage1}
\end{figure*}

Across all modes, the prompt enforces: (i) a strict dual-panel structure (never swapping front/back), (ii) a slightly oblique camera to expose richer local details than orthographic turnarounds, (iii) preservation of outfit colors and key accessories, and (iv) a pure background. Figure~\ref{fig:stage1} provides an example of Stage-1.

\subsection{Stage-2: Preview-to-Atlas via Fine-Tuned FLUX.2}

Stage-2 takes the Stage-1 dual-panel preview as conditioning input and generates a 512$\times$512 atlas image, designed to be a nearest-neighbor upscale of a 64$\times$64 Minecraft skin UV map. We fine-tune the FLUX.2 model on paired supervision constructed from the skin dataset. After generation, a deterministic decoder performs nearest-neighbor downsampling and converts the output into a \textbf{64$\times$64 RGBA} skin with a valid structure.

To improve stability and efficiency, we adopt \textbf{EvolveLoRA}, a progressive adapter-training curriculum that initializes each phase from the LoRA obtained in the previous phase: (i) text-to-image pre-adaptation with automatically constructed coarse prompts, (ii) image-to-image adaptation conditioned on front/back renders of the skin, and (iii) the final preview-to-atlas adaptation used in Stage-2.

\subsection{Why Two Stages?}

In principle, the most appealing solution is an end-to-end \emph{character$\rightarrow$skin atlas} generator. In practice, however, this approach is fragile for several reasons. First, there is no readily available large-scale dataset that pairs arbitrary character references with Minecraft skin atlases, making supervised training difficult. Second, real-world character inputs are highly non-canonical: they vary in camera angle, pose, resolution, cropping, and artistic style, and often contain substantial background clutter. This input noise makes it hard for a single model to learn a stable mapping while also satisfying strict UV-structure constraints.

A natural alternative is to synthesize training data by reversing the direction: starting from a skin preview, use a powerful MLLM to ``reconstruct'' a high-resolution character turnaround, and then train a model to map that turnaround back to the atlas. Unfortunately, Minecraft skins are already a heavily compressed representation. When asked to upsample them into detailed character sheets, generative models tend to hallucinate fine-grained textures, accessories, and shading cues that are not grounded in the original skin. These spurious details introduce label noise and inconsistency, which again destabilize subsequent fine-tuning.

BLOCK adopts a more pragmatic decomposition. Instead of using MLLMs to expand skins into high-fidelity character sheets for data construction, we use a powerful MLLM to perform the compression and normalization step: it maps arbitrary character references into a canonical Minecraft-style dual-panel preview with consistent framing, pose, and viewpoint. This preview retains the key semantic and stylistic attributes while removing nuisance factors. The remaining problem, \emph{preview$\rightarrow$atlas translation under strict UV validity}, is then handled by a specialized model trained with large-scale paired supervision. In short, BLOCK delegates semantic understanding and style commitment to the MLLM, and reserves pixel-perfect atlas validity for a dedicated translator.

\section{Fine-tuning Phase for Flux.2 Model (EvolveLoRA)}
\label{sec:fine-tune}

We construct all training pairs from \href{https://huggingface.co/datasets/nyuuzyou/Minecraft-Skins-20M}{\texttt{Minecraft-Skins-20M}}. Given limited computing resource, we do not perform full end-to-end fine-tuning on the hardest setting (3D-preview$\rightarrow$atlas) from scratch. Empirically, directly training on 3D-preview conditioning can be slow and unstable: the model must simultaneously learn the UV-structured mapping and handle the additional appearance variations introduced by preview rendering, which increases optimization difficulty.

To accelerate convergence under resource constraints, we adopt a three-phase curriculum (\textbf{EvolveLoRA}) that decomposes the task from easy to hard and progressively reuses the adapter from the previous phase. Concretely, we fine-tune with (i) text-to-image using automatically constructed coarse prompts derived from skin colors and part-wise patterns, (ii) image-to-image conditioned on front/back renders of the target skin, and (iii) the final preview-to-atlas training conditioned on Minecraft-style 3D dual-panel previews. Each phase uses roughly 10{,}000 training pairs, and each image is seen only once (one epoch), yielding a lightweight yet effective adaptation pipeline.

\subsection{Text-to-Image Adaptation}
\label{sec:t2i_adapt}

As the first phase of \textbf{EvolveLoRA}, we perform a lightweight text-to-image (T2I) adaptation on raw Minecraft skin UV atlases. The goal of this phase is not to solve the final preview-to-atlas translation task, but to quickly inject Minecraft-skin-specific priors into the model under limited compute. This provides a stable initialization for the subsequent image-conditioned phases.

Starting from raw $64\times 64$ RGBA skin PNGs, we convert each skin into an RGB training target and upscale it to $512\times 512$ using nearest-neighbor interpolation.
Concretely, we (i) fill transparent pixels with pure white, producing an RGB image with a clean background; and (ii) apply an integer $8\times$ upscale to match the training resolution used in later phases.

Since the skin dataset does not provide natural language descriptions, we generate captions automatically using a simple template-based procedure. For each skin, we sample several canonical UV regions (head, body, arms, legs, and optional hat/overlay) and compute the dominant color in each region by counting the most frequent RGB value.
We then map dominant RGB colors to a small vocabulary of named colors by nearest-color matching in RGB space (e.g., black/white/gray/red/blue).
Finally, we assemble a caption with a fixed structure:

\emph{``A Minecraft skin texture UV atlas, 64x64 pixel art layout. head is \dots, body is \dots, arms are \dots, legs are \dots.''}
If the hat/overlay region contains non-white pixels, we additionally append a phrase describing the overlay color (e.g., \emph{``has a blue hat overlay''}).

All captions explicitly emphasize pixel-art constraints (flat colors and hard edges), which encourages the model to avoid smooth shading and anti-aliasing artifacts. This phase is intentionally simple: it only requires paired (caption, target image) supervision and avoids the harder alignment problem introduced by preview-style conditioning.
Empirically, directly training on 3D preview inputs from scratch can be unstable because the model must simultaneously learn UV-structured placement and handle additional appearance variations from preview rendering. Indeed, it almost learn nothing from the very begining episode.
By first adapting the model to the skin domain with T2I, we obtain a robust LoRA initialization that accelerates convergence in the later front/back-conditioned image-to-image phase and the final preview-to-atlas phase.

\subsection{Image-to-Image Adaption}

As Phase II of \textbf{EvolveLoRA}, we perform image-to-image (I2I) adaptation where the conditioning input is a single turnaround image that already contains the same Minecraft character in front/back views, and the target is the corresponding \textbf{skin UV atlas}. This phase bridges the gap between pure text supervision (Phase I) and the final preview-conditioned translation (Phase III): the model learns to map a canonical two-view character depiction into a UV-structured atlas while remaining within the Minecraft domain, without yet introducing additional rendering variations from 3D preview images.

To obtain a consistent $512\times 512$ conditioning input without distortion, we apply a cover-and-center-crop transform: the turnaround image is resized (nearest-neighbor) to cover the target resolution and then center-cropped to exactly $512\times 512$. Compared with padding-based resizing, this avoids introducing artificial borders and keeps the character scale more consistent across the dataset.

Unlike Phase I, captions in Phase II are img2img-focused and do not describe part-wise colors. We use a small set of prompt templates that (i) explicitly state the reference contains front/back views of the same character, (ii) ask the model to output a \textbf{64$\times$64 UV atlas layout} (implemented as $512\times512$ training targets), and (iii) emphasize pixel-art constraints (flat colors, crisp edges, no blur/anti-aliasing). To improve robustness, we optionally randomize template phrasing with a deterministic seed. We also append the parsed \texttt{model\_type} (classic/slim) to the prompt to provide a lightweight conditioning signal for arm geometry.

\subsection{Preview-to-Atlas Adaption}
Phase III is the final step of \textbf{EvolveLoRA}, where we train the model to translate a \textbf{3D Minecraft preview} into the corresponding \textbf{skin UV atlas}. We synthesize paired training data by rendering preview images directly from each ground-truth skin. The preview rendering and the skin-to-preview image transformation are implemented using the open-source tool \href{https://github.com/Ghqqqq/mc-skin-to-sketch}{\texttt{mc-skin-to-sketch}}.

Given a raw $64\times64$ RGBA skin atlas, we render two oblique views of the same Minecraft character:
(i) a front view and (ii) a back view, using a lightweight cube-based renderer that respects both the base layer and the outer/overlay layer. The default camera matches our Stage-1 specification (yaw $\approx 24^\circ$, pitch $\approx 30^\circ$) to minimize the domain gap between synthetic previews and MLLM previews. Each rendered view is placed into a fixed dual-panel layout (front on the left, back on the right) with a small gap and a pure white background.

To improve robustness to inevitable viewpoint variations from MLLM-generated previews, we apply camera jitter with a controlled probability. With probability $1-p$ (default $p=0.7$ keep-prob), we perturb yaw and pitch independently by a random offset within $\pm \Delta$ degrees (default $\Delta=10^\circ$). This augmentation encourages the model to learn a stable mapping from preview appearance to UV structure under small camera changes.

The training target is the corresponding $512\times512$ atlas image obtained by (i) compositing RGBA onto a white background and (ii) applying an $8\times$ nearest-neighbor upscale from $64\times64$ to $512\times512$. We use $512\times512$ targets to align with the image-generation backbone resolution and to keep a consistent format across phases. Prompts in Phase III explicitly state that the reference image is a 3D Minecraft character with front/back views in one image and ask for the corresponding $64\times64$ UV atlas layout. 

\section{Qualitative Examples}
\label{sec:examples}

We present qualitative results of \textbf{BLOCK}. Each example group contains three images: (i) the input character concept/reference, (ii) the Stage-1 Minecraft-style dual-panel preview, and (iii) the final Stage-2 generated skin atlas (shown as a $512\times512$ atlas or a rendered preview).

\begin{figure*}[ht]
\centering
\begin{subfigure}[t]{0.32\textwidth}
  \centering
  \includegraphics[width=\linewidth]{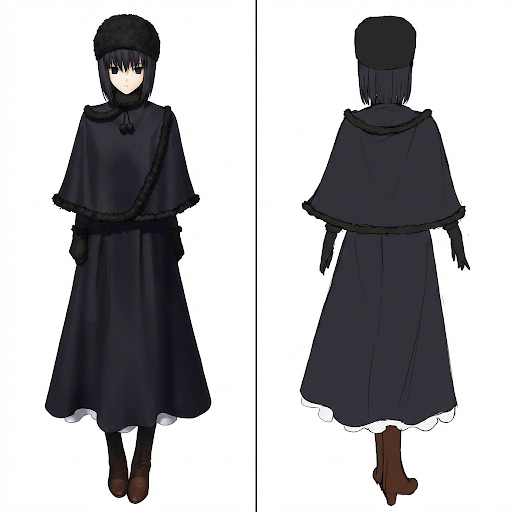}
  \caption{Character reference}
\end{subfigure}\hfill
\begin{subfigure}[t]{0.32\textwidth}
  \centering
  \includegraphics[width=\linewidth]{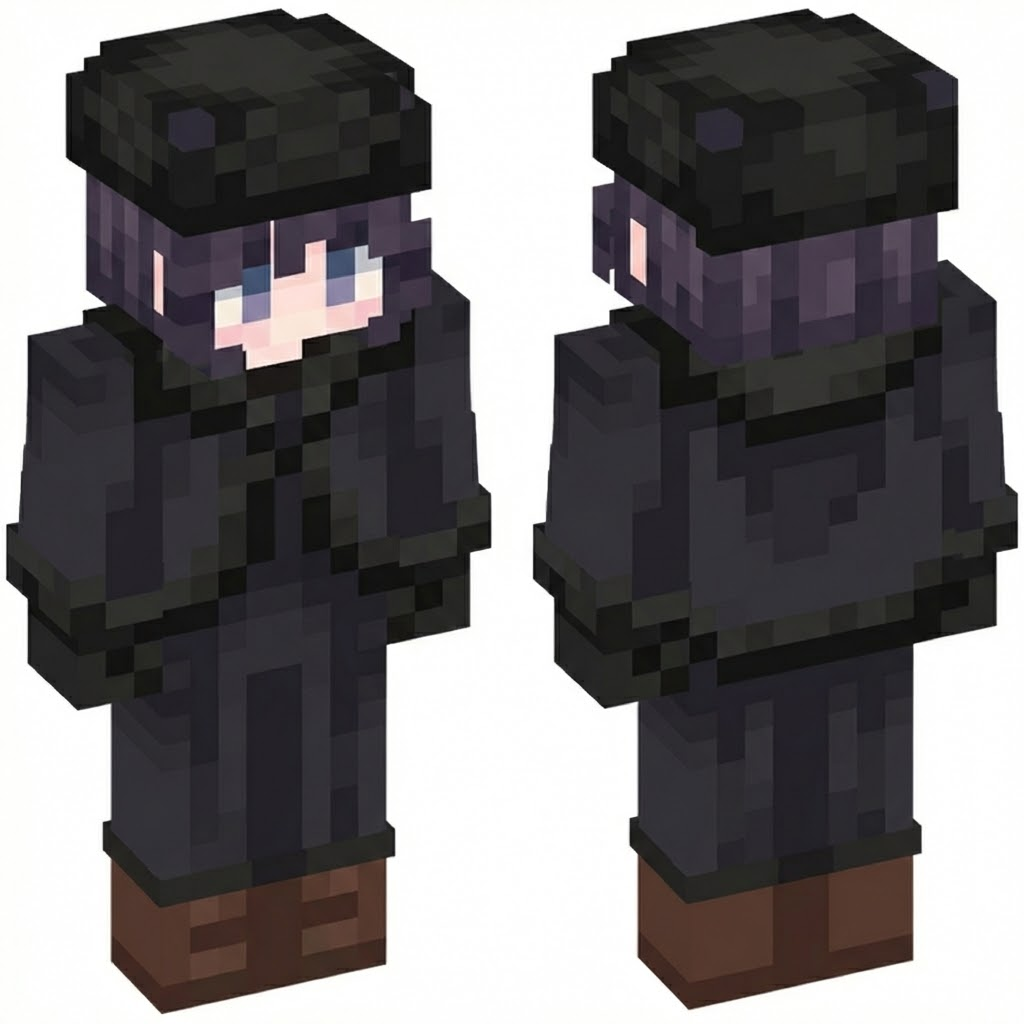}
  \caption{Stage-1 preview}
\end{subfigure}\hfill
\begin{subfigure}[t]{0.32\textwidth}
  \centering
  \includegraphics[width=\linewidth]{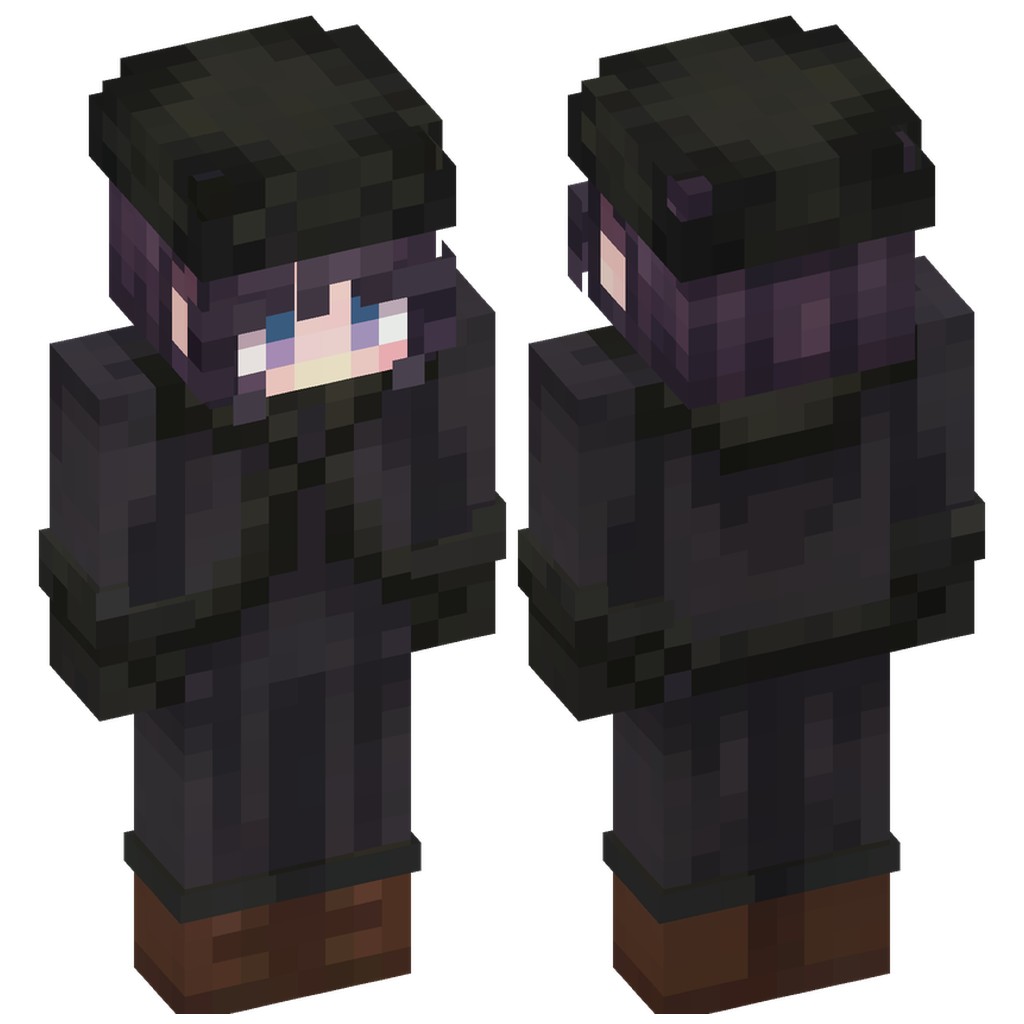}
  \caption{Final skin preview}
\end{subfigure}
\caption{Example 1, A beginning scenario.}
\label{fig:ex_group1}
\end{figure*}

\begin{figure*}[ht]
\centering
\begin{subfigure}[t]{0.32\textwidth}
  \centering
  \includegraphics[width=\linewidth]{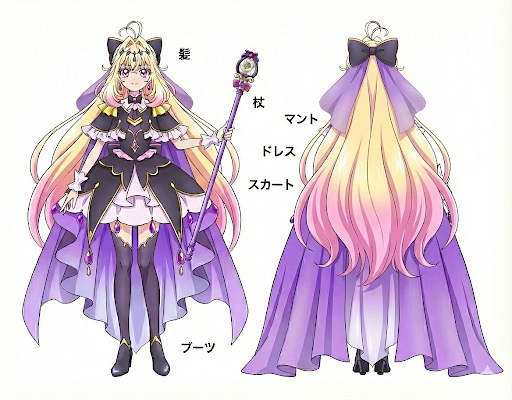}
  \caption{Character reference}
\end{subfigure}\hfill
\begin{subfigure}[t]{0.32\textwidth}
  \centering
  \includegraphics[width=\linewidth]{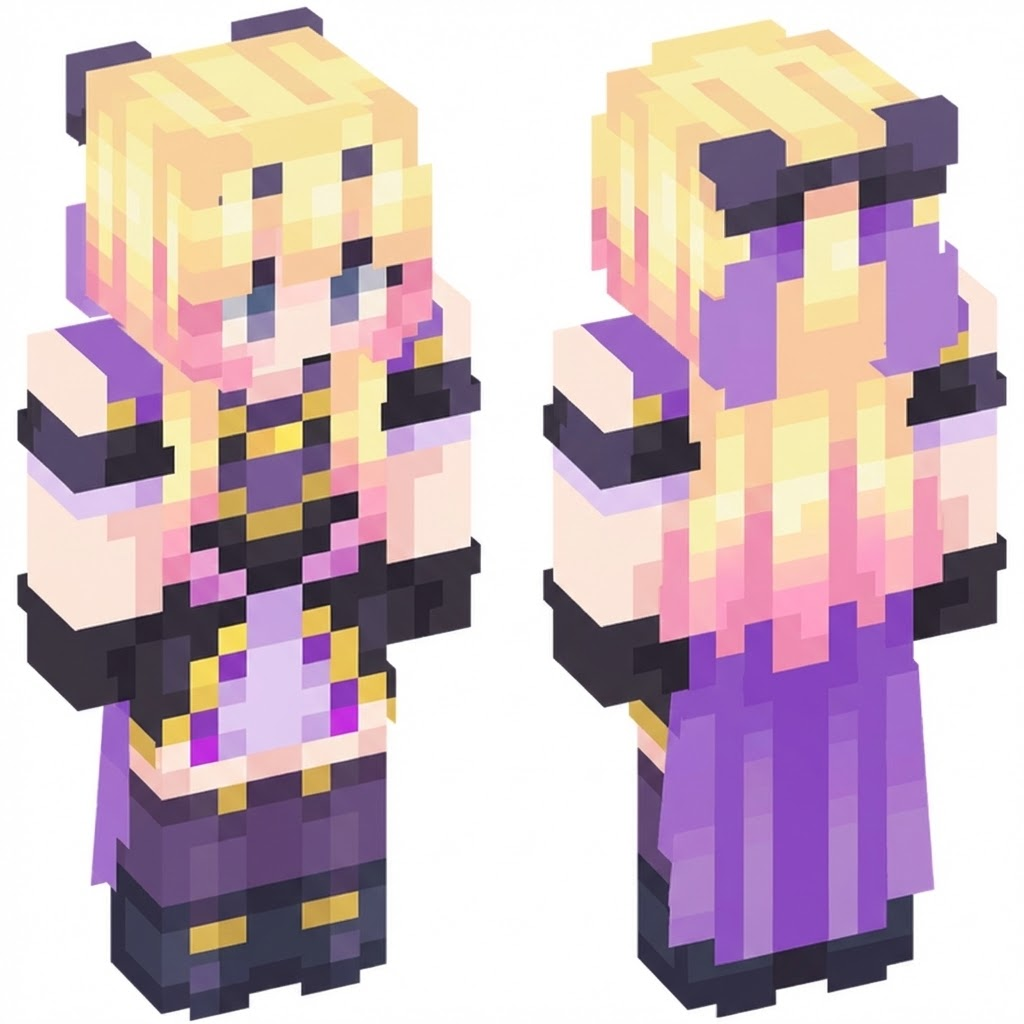}
  \caption{Stage-1 preview}
\end{subfigure}\hfill
\begin{subfigure}[t]{0.32\textwidth}
  \centering
  \includegraphics[width=\linewidth]{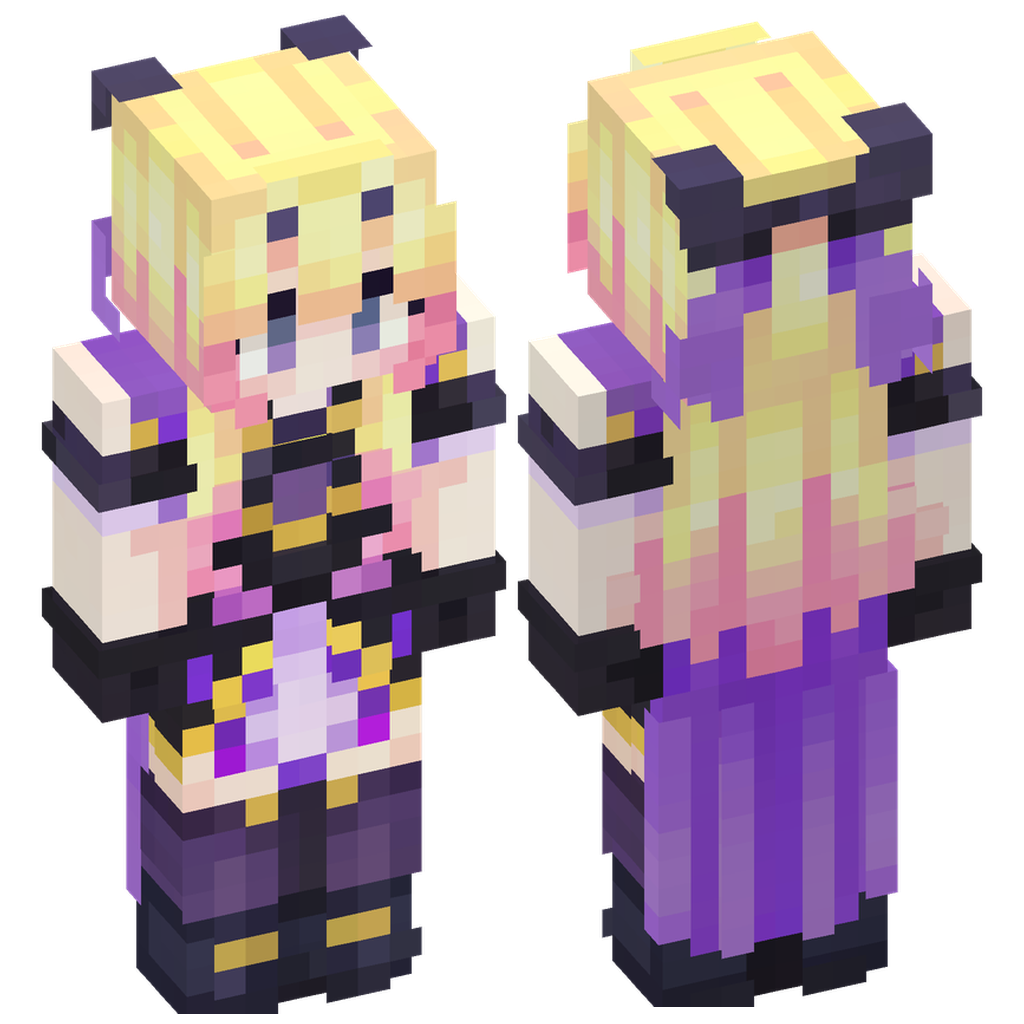}
  \caption{Final skin preview}
\end{subfigure}
\caption{Example 2, A considerably more difficult scenario involving many fine-grained details.}
\label{fig:ex_group2}
\end{figure*}

\begin{figure*}[ht]
\centering
\begin{subfigure}[t]{0.32\textwidth}
  \centering
  \includegraphics[width=\linewidth]{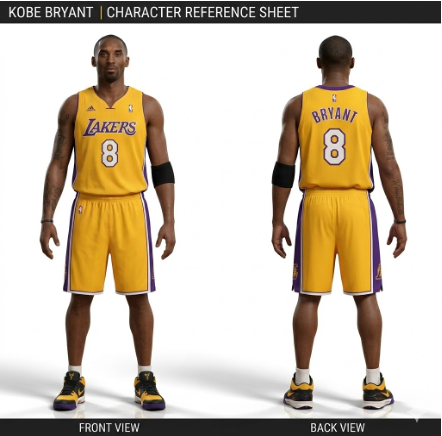}
  \caption{Character reference}
\end{subfigure}\hfill
\begin{subfigure}[t]{0.32\textwidth}
  \centering
  \includegraphics[width=\linewidth]{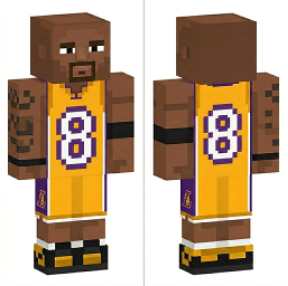}
  \caption{Stage-1 preview}
\end{subfigure}\hfill
\begin{subfigure}[t]{0.32\textwidth}
  \centering
  \includegraphics[width=\linewidth]{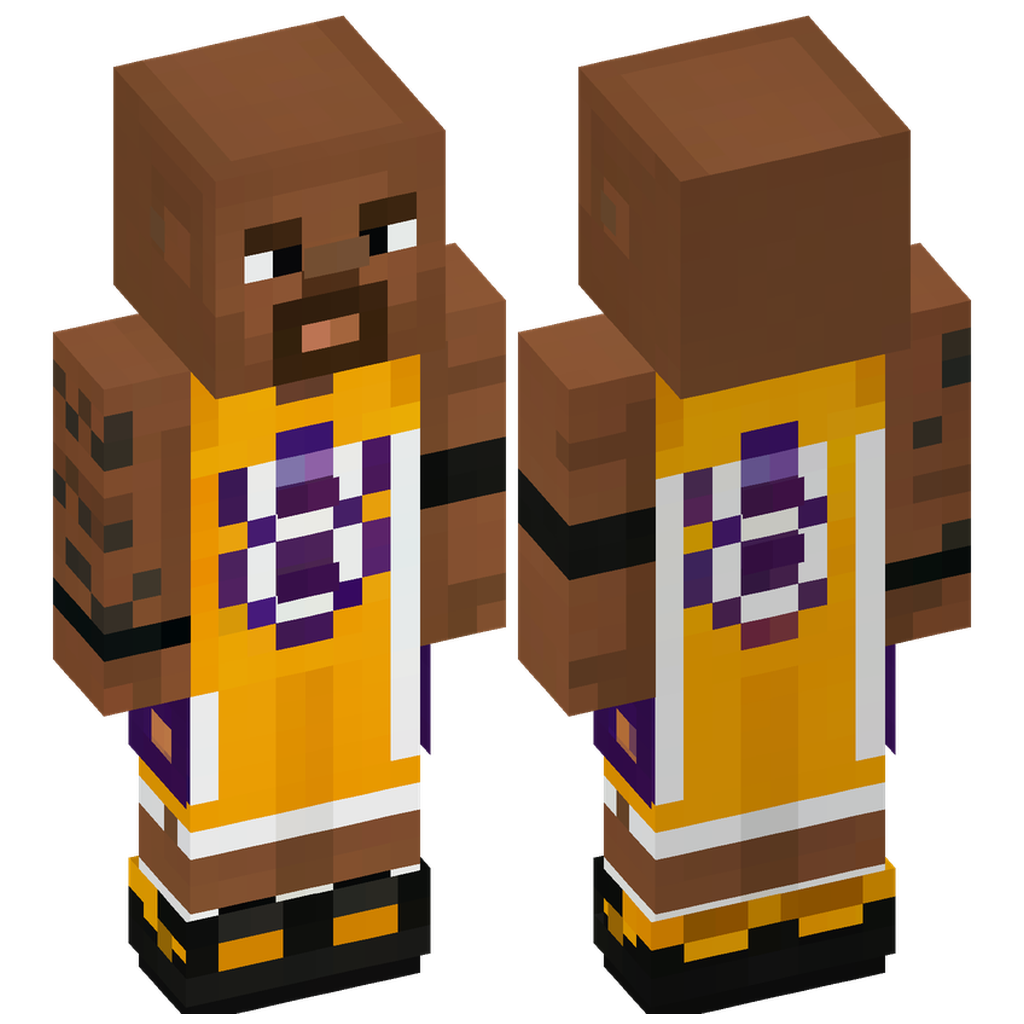}
  \caption{Final skin preview}
\end{subfigure}
\caption{Example 3, The most difficult case study, where the preview from Stage-1 fails to correctly compress the details from character concept.}
\label{fig:ex_group3}
\end{figure*}



\section{Limitations and Future Work}

\paragraph{Downsample-Friendly Previews.}
In fact, Figure~\ref{fig:pipeline} can be viewed as a \emph{bad-case} example. It highlights a failure mode where the MLLM does not compress visual details aggressively enough when producing the Stage-1 preview. In such cases, the Stage-2 translator largely succeeds at mapping these elements to plausible UV locations, but it does not further ``compress'' or simplify them into a representation that survives discrete decoding. Consequently, when we apply the deterministic downsampling step to obtain the $64\times64$ skin, many fine details are lost or aliased, leading to a visually unsatisfactory final skin. Future work may incorporate fine-tuning process that let Stage-2 model learn how to compress details

\paragraph{Direct view-to-skin without Stage-1.}
Our long-term goal is to generate skins directly from arbitrary multi-view inputs (or even a single view) without relying on Stage-1 synthesis. In the current report, Stage-1 improves controllability and standardizes the conditioning distribution, but introduces dependence on an external MLLM.

\paragraph{Data coverage and rare details.}
Some failures arise from limited coverage of rare accessories, hairstyles, or face details. And Stage-1 does not always directly output ideal result, which might require multiple rollouts.

\paragraph{Open-source vs proprietary components.}
While Stage-2 is open-source, Stage-1 depends on an MLLM. We provide prompt templates and recommend using any capable multimodal model; replacing Stage-1 with a fine-tuned open model is an important direction.

\paragraph{Compute and deployment cost.}
Our current Stage-2 fine-tuning is based on FLUX.2 9B, which can be challenging to deploy on commodity GPUs with limited VRAM. We believe the preview-to-atlas translation task does not intrinsically require such a large base model. A promising direction is to distill or transfer the learned capability to smaller backbones (e.g., 4B-class models), enabling wider adoption and faster iteration for open-source users.

\paragraph{Datasets and benchmarks.}
Finally, we advocate for the development of high-quality Minecraft skin datasets and standardized benchmarks. Progress on pixel-perfect skin generation is currently hindered by the lack of widely accepted evaluation protocols, curated test suites, and metrics that jointly measure UV-structure correctness, overlay validity, and character-level consistency across views. A community-driven effort to build cleaner datasets, define challenging and diverse benchmarks (including rare accessories and edge cases), and provide reproducible evaluation tooling would greatly accelerate research and open-source development in this space.

\bibliographystyle{unsrtnat}
\bibliography{references}  
\newpage
\appendix

\section{Reference Figure}\label{app:refig}
\begin{figure}[H]
\centering
\begin{minipage}[t]{0.45\linewidth}
  \centering
  \includegraphics[width=\linewidth]{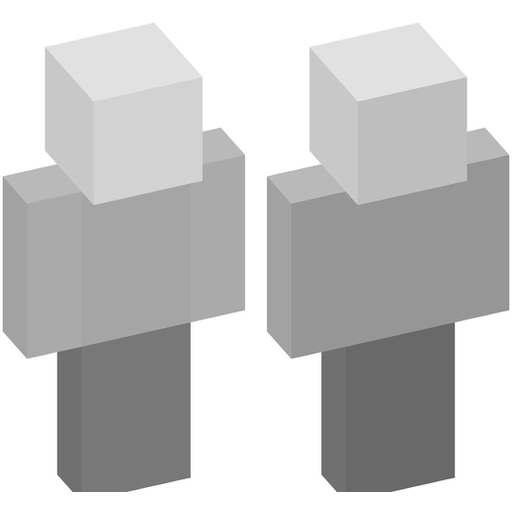}
\end{minipage}\hfill
\begin{minipage}[t]{0.45\linewidth}
  \centering
  \includegraphics[width=\linewidth]{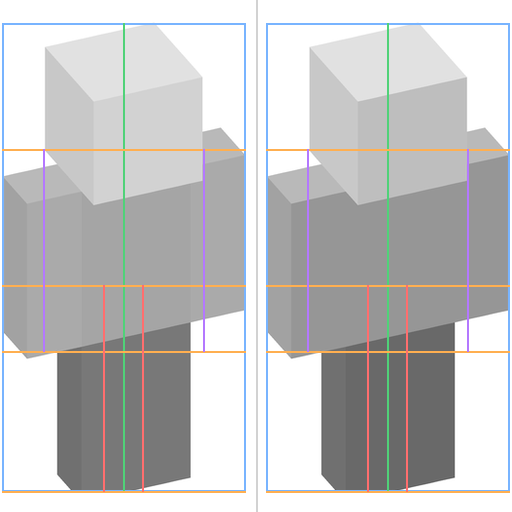}
\end{minipage}
\caption{Layout reference (left) and pose anchor (right).}
\label{fig:layout_pose_refs}
\end{figure}
\section{Prompts}\label{app:prompts}

\begin{promptbox}[Stgae-1 Mode-I Full Prompt (Gemini Nano Banana Pro)]
\ttfamily\scriptsize
You are a Minecraft skin creator. You receive 3 input images:

INPUTS (3 images):
[A] Character reference: same anime character with FRONT + BACK views (side-by-side or top/bottom).
[B] Layout reference: target Minecraft dual-panel framing and camera.
[C] Pose-anchor reference: strict limb-position guide.

OUTPUT:
Generate exactly ONE final image only.
- Pure white background \#FFFFFF.
- Two panels only, side-by-side:
  - Left = FRONT
  - Right = BACK
- Never swap.
- No extra views, no extra images, no text, no watermark, no UI, no props, no scene.

VIEW PARSING (HARD):
- If [A] is side-by-side: default left=front, right=back (unless explicit text labels say otherwise).
- If [A] is top/bottom: rearrange to left=front, right=back.
- Left panel uses front-view info only.
- Right panel uses back-view info only.
- Do not copy front-only details to back or back-only details to front.

CAMERA / ORIENTATION (LOCKED):
- Near-orthographic look (very weak perspective).
- Front panel: +24° yaw (front-right), +30° downward tilt.
- Back panel: 180°+24° yaw (back-right), same +30° downward tilt.
- Front/back orientation relationship must remain parallel in 3D space.
- Match framing style of [B].

POSE LOCK (ULTRA STRICT):
- Minecraft default standing pose only.
- Head, torso, arms, legs are clean cuboids with straight vertical sides.
- Arms:
  - Start at shoulder line, hang straight down.
  - No bend, no forward/back swing, no outward spread.
  - Left/right symmetry required.
- Legs:
  - Start at hip line, go straight down.
  - Parallel, no crossing, no toe-in/toe-out, no bend.
- Feet bottoms must share the same baseline.
- Limb-position tolerance:
  - Vertical drift of each arm centerline from shoulder to hand <= 1 px.
  - Vertical drift of each leg centerline from hip to foot <= 1 px.
  - Left/right symmetry error <= 2 px.
  - Foot baseline mismatch <= 1 px.
- If any tolerance fails, regenerate.

PROPORTION LOCK (CLASSIC):
- Approx height ratio head:torso:legs = 8:12:12.
- Arm height about 12, arm width about 4; leg width about 4 (relative block proportions).
- Same scale and body proportions in both panels.
- Full body visible (head to feet), feet aligned, minimal empty margin, no cropping.

PIXEL FORMAT LOCK (VERY HARD):
- True pixel art only. No smooth high-precision rendering.
- Build on a low-resolution pixel grid first, then upscale by integer factor using NEAREST only.
- No anti-aliasing, no blur, no feathering, no denoise smoothing.
- No sub-pixel edges, no fractional blending, no soft gradients.
- Use hard square pixel blocks; staircase edges are acceptable/preferred.
- Palette limit:
  - Total character palette recommended <= 48 colors (excluding white background).
  - Hair palette <= 6 colors.
- Shading limit: max 2-3 tones per material.
- No isolated 1-pixel noise clusters.

HAIR PIXELIZATION LOCK (CRITICAL):
- Hair must be converted into blocky pixel masses, not smooth strands.
- No thin strand rendering, no wispy/soft edges, no semi-transparent flyaways.
- Minimum visible hair feature width: >= 2 px at the working pixel grid.
- Hairline, bangs, side locks, and back hair must be represented as square block clusters.
- If hair appears too smooth or too detailed, re-pixelate and regenerate.

FIDELITY:
- Preserve hairstyle structure, outfit structure, color-block layout, patterns, accessories.
- Front details from front reference; back details from back reference.
- If back is unclear, do minimal consistent completion only; do not invent new design elements.

FINAL SELF-CHECK (MUST PASS ALL):
1) Left=FRONT, Right=BACK correct.
2) Arms vertical, symmetric, and anchored correctly.
3) Legs vertical, parallel, and feet aligned.
4) Same character / same outfit in both panels.
5) Pure white background, one dual-panel image only.
6) Hair is clearly blocky pixel mass (not smooth high-precision strands).
If any check fails, regenerate before final output.
\end{promptbox}

\begin{promptbox}[Stage-2 Mode-II Full Prompt with Style Reference  (Recommended)]
\ttfamily\scriptsize
You are a Minecraft skin creator. You receive 4 input images:
[A] Character reference: same anime character with FRONT + BACK views (side-by-side or top/bottom).
[B] Layout reference: target Minecraft dual-panel framing and camera.
[C] Pose-anchor reference: strict limb-position guide.
[D] Style reference: target Minecraft skin-preview rendering style (pixel granularity, palette, shading levels, blocky hair treatment).

Task:
Convert [A] into a Minecraft player block-character preview.
Output exactly ONE final image only.
Priority order (highest to lowest):
1) Hard output structure and front/back order.
2) Hard pose/limb placement/proportion constraints.
3) Character fidelity from A.
4) Rendering style consistency with D (without copying D identity).

Hard output constraints:
- Pure white background \#FFFFFF.
- Exactly two side-by-side panels: Left=FRONT, Right=BACK.
- Never swap front/back.
- No extra views, no extra images, no text, no watermark, no UI, no props, no scene.

View parsing:
- If A is side-by-side: default left=front, right=back (unless explicit labels say otherwise).
- If A is top/bottom: rearrange to left=front, right=back.
- Left panel uses front-view info only; right panel uses back-view info only.
- No cross-copying of view-specific details.

Camera/orientation lock:
- Near-orthographic look (very weak perspective).
- Front panel: +24° yaw (front-right) and +30° downward tilt.
- Back panel: 180°+24° yaw (back-right) and the same +30° downward tilt.
- Front/back orientation relationship must remain parallel in 3D space.
- Match framing from B and limb anchoring from C.

Pose lock (ultra strict):
- Minecraft default standing pose only; head/torso/arms/legs are clean cuboids.
- Arms drop vertically from shoulder line; no bend, no forward/back swing, no outward spread; left/right symmetric.
- Legs drop vertically from hip line; parallel; no crossing, no toe-in/out, no bending.
- Both foot bottoms must be on the same baseline.
- Tolerances: arm centerline drift <=1 px, leg centerline drift <=1 px, left/right symmetry error <=2 px, foot baseline mismatch <=1 px.
- If any tolerance fails, regenerate.

Proportion lock (classic model):
- Approx height ratio head:torso:legs = 8:12:12.
- Arm height about 12 and width about 4; leg width about 4 (relative block proportions).
- Same scale/proportions in both panels.
- Full body visible head-to-feet; minimal empty margin; no cropping.

Style lock from D (hard):
- Learn rendering style from D only; do NOT copy D's character identity, outfit motifs, or exact design.
- True pixel-art pipeline: construct on low-res grid first, then integer upscale with NEAREST.
- No anti-aliasing, no blur, no feathering, no denoise smoothing, no sub-pixel blending, no soft gradients.
- Hard square pixel boundaries required; staircase edges are acceptable.
- Recommended palette: <=48 character colors (excluding white background); hair palette <=6; 2-3 shading steps per material.
- No isolated 1-pixel noise clusters.

Hair pixelization lock (critical):
- Hair must be blocky pixel masses, not smooth strands.
- No thin wispy strands, no soft edges, no semi-transparent flyaways.
- Minimum visible hair feature width >=2 px at working pixel grid.
- Hairline, bangs, side locks, and back hair must be represented as square pixel clusters.
- If hair still looks high-precision/smooth, re-pixelate and regenerate.

Fidelity from A:
- Preserve hairstyle structure, outfit structure, color-block layout, pattern placement, and accessories.
- Front details come from A-front; back details come from A-back.
- If back is unclear, do minimal consistent completion only; do not invent new design elements.

Final self-check (must pass all):
1) Left=FRONT and Right=BACK.
2) Limbs satisfy C anchors and tolerance limits.
3) Legs are parallel and feet baselines are aligned.
4) Same character and same outfit in both panels.
5) Pure white background and exactly one dual-panel image.
6) Style matches D (pixel blocks, limited palette, low-step shading).
7) D identity/design is not copied.
8) Hair is clearly blocky pixel mass.
If any check fails, regenerate before final output.
\end{promptbox}

\begin{promptlisting}{Stage-2 Mode-III Meta-Prompt Injection (Strongly Recommended)}
You are a Minecraft skin preview renderer. Your top priority is to match the preview render STYLE of Image C exactly. Create a cute Minecraft skin preview.

INPUT IMAGES (3, in order):

Image A = character design ref (front/back). Use ONLY for clothing, hair, colors, and character details.

Image B = camera / pose / framing + Minecraft geometry lock (front/back mannequin). Use as STRICT template for camera, proportions, placement, body orientation, and overall panel layout.

Image C = STYLE MASTER reference (front/back preview). This is the render-style ground truth.

GOAL

Generate ONE single image containing two Minecraft preview renders:
LEFT = FRONT view
RIGHT = BACK view

Requirements:
- Geometry, pose, camera, framing, panel layout, and placement must match Image B exactly.
- Render style must match Image C exactly.
- Skin texture / character design must follow Image A only.

This is a template-transfer task, not a redesign task.
Do not reinterpret the composition.
Do not redesign the renderer.
Do not change the body model to better fit the outfit.
A only controls texture content; B controls structure; C controls render style.

PRIORITY ORDER (ABSOLUTE)

If anything conflicts, follow this order:
(1) Image B = camera / pose / proportions / framing / panel placement must be identical
(2) Image C = render style must be identical
(3) Image A = skin design details

Within ``style,'' Image C has highest authority:
lighting, shading, anti-aliasing, edge treatment, pixel crispness, material response, contrast, saturation, outline behavior, and the overall voxel-preview feeling must match Image C exactly.

HARD OUTPUT REQUIREMENTS (MUST FOLLOW)

- Output exactly ONE image only.
- Two panels only: LEFT = FRONT, RIGHT = BACK.
- Match Image B exactly:
  same yaw, same pitch, same zoom, same placement, same body orientation, same limb thickness, same head size, same margins, same spacing, same front/back arrangement.
- Match Image C exactly in rendering style:
  same voxel shading softness / hardness,
  same edge AA / crispness,
  same highlight and shadow logic,
  same pixel-texture readability,
  same contrast / saturation response,
  same cloth / skin / shoe material rendering behavior,
  same overall Minecraft preview look.
- Background must be pure white (#FFFFFF) only.
- Do NOT output a UV atlas.
- Output must be a rendered Minecraft skin preview like Image C, while using Image B's exact pose/framing/layout.

SKIN DESIGN CONTENT (from Image A, translated into Minecraft skin)

Create a Minecraft skin that matches Image A precisely:

Character identity and palette:
- [CHARACTER_TYPE]
- hair color: [HAIR_COLOR]
- eye color: [EYE_COLOR]
- outfit palette: [OUTFIT_PALETTE]
- overall feeling: [OVERALL_FEELING]

Hair:
- [HAIR_LENGTH_AND_BASE_SHAPE]
- [BANGS_DESCRIPTION]
- [SIDE_LOCKS_OR_FACE_FRAMING]
- [HAIR_ENDS_DESCRIPTION]
- [AHOGE_OR_TOP_HAIR_DESCRIPTION]
- [HEAD_ACCESSORY_DESCRIPTION]
- [HEAD_ACCESSORY_POSITION_CONSTRAINT]
- [FRONT_BACK_HAIR_CONSISTENCY]

Face:
- [EYE_DESCRIPTION]
- simple Minecraft pixel face style, not anime facial proportions
- [FACE_EXPRESSION_DESCRIPTION]

Outfit:
- [UPPER_GARMENT_DESCRIPTION]
- [COLLAR_OR_NECKLINE_DESCRIPTION]
- [CHEST_ORNAMENT_DESCRIPTION]
- [SLEEVE_AND_CUFF_DESCRIPTION]
- [MAIN_SKIRT_OR_DRESS_DESCRIPTION]
- [VISIBLE_LAYERING_OR_HEM_DESCRIPTION]

Back details:
- [BACK_VIEW_MAIN_DESCRIPTION]
- [BACK_ACCESSORY_DESCRIPTION]
- [BACK_SIMPLIFICATION_OR_CONSERVATIVE_RULE]

Legwear / shoes:
- [LEGWEAR_DESCRIPTION]
- [SHOE_DESCRIPTION]
- [SHOE_CONSTRAINT]

CONSISTENCY REQUIREMENTS

- Keep front and back outfit details logically consistent.
- Keep [KEY_CONSISTENCY_ITEMS] coherent between both views.
- Preserve the character's [KEY_COLOR_IDENTITY].
- The result must still read immediately as a Minecraft skin preview, not an anime illustration.

STYLE ENFORCEMENT (VERY IMPORTANT)

Treat Image C as the exact style template to imitate.

The final output should look like:
``Image C's renderer and post-processing''
applied to
``Image B's exact camera, exact model geometry, exact framing, and exact panel layout''
with
``Image A's character texture and outfit design.''

Do NOT drift into another Minecraft render look.
Do NOT stylize beyond Image C.
Do NOT reinterpret the character freely.
Do NOT change the silhouette, framing, or preview format to better express the outfit.

NEGATIVE PROMPT

different render style than Image C, different shading model, different AA style, painterly, smooth 2D anime illustration, photorealism, cinematic lighting, glossy realistic skin, different voxel renderer, different outline style, different contrast or saturation than Image C, different camera than Image B, different pose, different proportions, non-Minecraft body, bending limbs, long legs, slim waist exaggeration, chibi proportions, extra accessories, text, watermark, UI, background scene, floor, platform, wrong panel order, wrong front/back arrangement, wrong framing, wrong zoom, wrong placement, wrong margins, wrong body scale, wrong Minecraft model proportions, non-Minecraft render, weak voxel texture readability, [CHARACTER_SPECIFIC_NEGATIVE_ITEMS]
\end{promptlisting}

\begin{promptlisting}{Stage-2 Mode-III Character-to-template compiler prompt}
You are a character-to-template prompt compiler.

I will give you:
- Image A = a character design reference image
- a fixed Minecraft render prompt skeleton with placeholders

Your task is to inspect Image A and fill in the character-related placeholders in the prompt skeleton.

This is NOT a rewrite task for the whole prompt.
This is a placeholder-filling task.

Follow these rules strictly:

1. Preserve all non-character control text
You must keep the following sections structurally unchanged and semantically strong:
- INPUT IMAGES
- GOAL
- PRIORITY ORDER
- HARD OUTPUT REQUIREMENTS
- STYLE ENFORCEMENT
- all B/C control wording
- all geometry / framing / render-style constraints
- all generic negative constraints

Do not weaken them.
Do not paraphrase them unless necessary.
Do not make them softer.

2. Only fill the character-related placeholders
You may only replace placeholders inside:
- Character identity and palette
- Hair
- Face
- Outfit
- Back details
- Legwear / shoes
- KEY_CONSISTENCY_ITEMS
- KEY_COLOR_IDENTITY
- CHARACTER_SPECIFIC_NEGATIVE_ITEMS

3. Be conservative and visual
Describe only what is visible or strongly implied by Image A.
Prefer:
- color
- placement
- length
- layer structure
- collar shape
- sleeve shape
- skirt structure
- visible back details
- legwear
- shoes
Avoid over-interpreting abstract fashion labels unless visually obvious.

4. Do not invent missing details
If some back detail is unclear, use conservative wording such as:
- keep back details subtle and consistent with Image A
- do not invent oversized rear accessories

5. Accessory positions must be explicit
For any bow, ribbon, clip, ornament, pendant, ahoge, or visible accessory, describe its location precisely:
- front-centered / side-mounted / back-of-head / rear crown / top-center / near one temple
Also include wrong placements in CHARACTER_SPECIFIC_NEGATIVE_ITEMS.

6. Keep B > C > A logic intact
If anything in Image A seems incompatible with the mannequin proportions, camera, or render style, do NOT modify B or C instructions.
Only express the character through texture and visible design translation.

7. Output format
Output exactly two parts:

Part 1: Filled placeholder summary
List the final values you inferred for each placeholder.

Part 2: Final completed prompt
Return the fully completed prompt with all placeholders replaced.
Output the final completed prompt in one plain text code block only.
Do not include any extra explanation outside these two parts.

Now inspect Image A and fill the prompt skeleton I provide below.
\end{promptlisting}






\end{document}